\documentclass[journal]{IEEEtran}

\ifCLASSOPTIONcompsoc
   \usepackage[nocompress]{cite}
\else
   \usepackage{cite}
\fi

\ifCLASSINFOpdf
  \usepackage[pdftex]{graphicx}

\else

\fi

\usepackage{booktabs}
\usepackage{multirow}
\usepackage{amsfonts}   
\usepackage{threeparttable}
\usepackage{amsmath}
\usepackage{array}

\usepackage{mdwmath}
\usepackage{mdwtab, color}

\usepackage{orcidlink} 

\begin{document}

\title{ A Real-Time Multi-Task Learning System for Joint  Detection of Face, Facial Landmark  and Head Pose}

\author{
        Qingtian~Wu$^{\orcidlink{0000-0002-2475-9156}}$,~\IEEEmembership{Member,~IEEE,} 
        Xiaoming~Wang$^{\orcidlink{0000-0001-5118-4581}}$,~\IEEEmembership{Student Member, IEEE,}  \\
        Liming~Zhang$^{\orcidlink{0000-0002-2664-8193}}$,~\IEEEmembership{Member,~IEEE,}  and   
        Fei~Richard Yu$^{\orcidlink{0000-0003-1006-7594}}$,~\IEEEmembership{Fellow,~IEEE}  
\IEEEcompsocitemizethanks{
\IEEEcompsocthanksitem Qingtian Wu,   Xiaoming Wang and Liming~Zhang (Corresponding author) 
 are with the Department of Computer and Information Science, Faculty of Sciences and Technology, University of Macau 
 (E-mail: \{yc07452, lmzhang\}@um.edu.mo).
\protect
\IEEEcompsocthanksitem  
F.~Richard Yu   is with the Shenzhen Key Laboratory of Digital and Intelligent Technologies and Systems, Shenzhen University, Shenzhen 518060, China (e-mail: yufei@szu.edu.cn).
\IEEEcompsocthanksitem 
This work was supported in part by the Science and Technology Development Fund of Macau SAR under Grant 0060/2021/A and 0071/2022/A, and in part by the Multi-Year Research Grant under Grant MYRG2022-00193-FST.
 \protect
}
}

\markboth{Script for the IEEE Internet of Things Journal,~Vol.~**, No.~**, **~2023}%
{Qingtian \MakeLowercase{\textit{et al.}}: Multi-Task Learning}
%

\IEEEtitleabstractindextext{

\begin{abstract}


  Extreme head postures pose a common challenge across a spectrum of facial analysis tasks, including face detection, facial landmark detection (FLD), and head pose estimation (HPE). 
  These   tasks are interdependent, where accurate FLD relies on robust face detection, and HPE is intricately associated with these key points.
  This paper focuses on the integration of these tasks, particularly when addressing the complexities posed by large-angle face poses. 
  The primary contribution of this study is the proposal of a real-time multi-task detection system capable of simultaneously performing joint detection of faces, facial landmarks, and head poses. 
  This system builds upon the widely adopted YOLOv8 detection framework. It extends the original object detection head by incorporating additional landmark regression head, enabling efficient localization of crucial facial landmarks. 
  Furthermore, we conduct  optimizations and enhancements on various modules within the original YOLOv8 framework. 
  To validate the effectiveness and real-time performance of our proposed model, we conduct  extensive experiments on 300W-LP and AFLW2000-3D datasets. The results obtained verify the capability of our model to tackle large-angle face pose challenges while delivering real-time performance across these interconnected tasks.


\end{abstract}

}

\maketitle

\IEEEdisplaynontitleabstractindextext

\IEEEpeerreviewmaketitle

\section{Introduction}
\label{YOLOHeadPose:introduction}

\IEEEPARstart{I}{n} recent years, the rapid advancement of deep learning techniques has revolutionized various computer vision tasks, particularly in the tasks of facial analysis tasks, including  face detection, facial landmark detection (FLD), and head pose estimation (HPE). Among them, face detection is a continuously and widely discussed topic;
  FLD  entails identifying key points on a face, such as eyes, nose, and mouth corners; 
   HPE  involves determining the orientation and position of a person's head relative to the camera or a reference frame. 
 These tasks have witnessed substantial progress and development, making them essential in numerous applications, including human-computer interaction  \cite{9881556}, face recognition \cite{9225055}, facial expression analysis \cite{9352023} and  driving assistance system \cite{9467376, 9695956}.

\begin{figure}[t]
  \begin{center}
      \includegraphics[width=1 \linewidth]{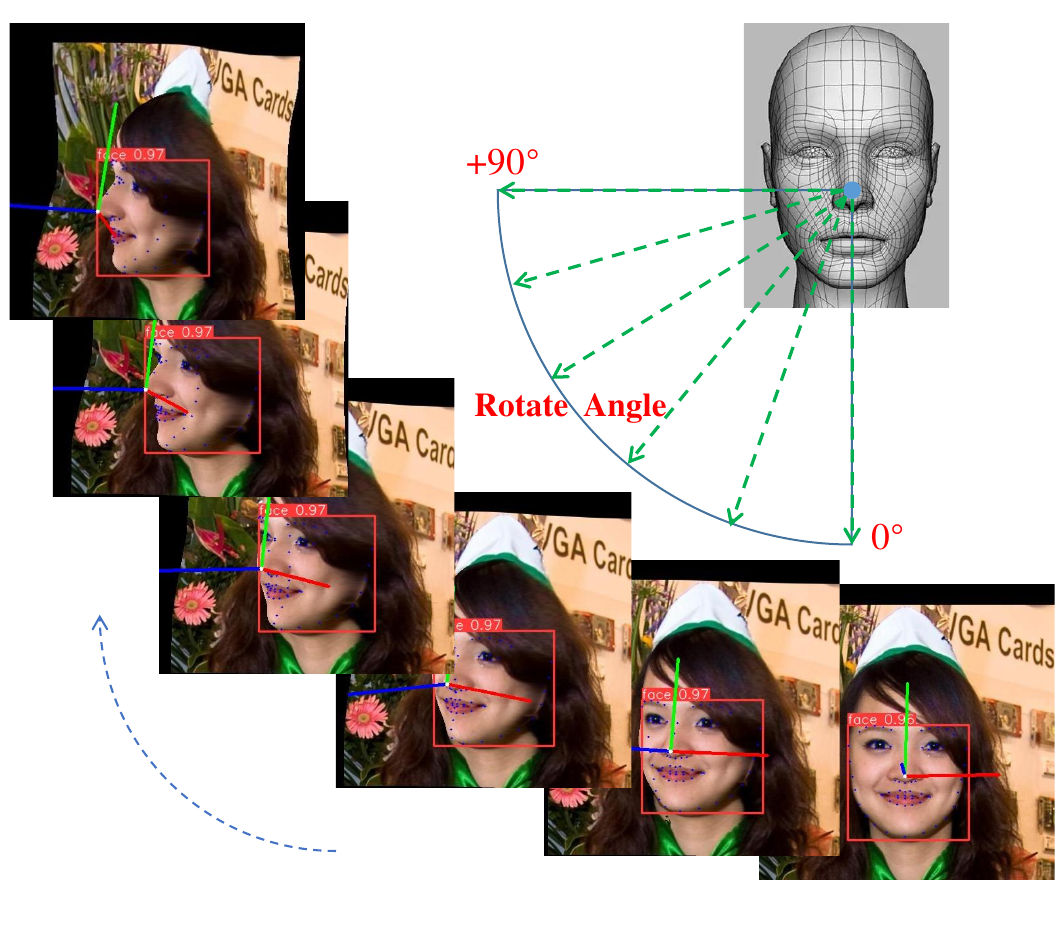}\\
      \caption{
        Our multitask learning system can simultaneously detect faces, facial landmarks, and head posture, from front to side faces.      (Best view in zoom).
      }
      \label{ParadigmCompare}
  \end{center}  
\end{figure}

However, despite the promising results achieved in each individual tasks, at least two challenges hinder their seamless integration into internet of things (IoT) applications:
\begin{itemize} 
  \item \noindent \textbf{Real-time requisites of practical systems.} 
  For instance, this timely response plays a pivotal role in averting hazardous driving behaviors, ultimately safeguarding lives and mitigating potential accidents in assisted driving systems. 
  However, the conventional approach \cite{guo2020towards}
   involves a sequential procedure: first detecting faces, followed by key point detection, and subsequently assessing fatigue status or ascertaining head posture based on these outputs. 
  Yet, this sequential cascade not only consumes time but also exerts a substantial demand on computational resources. 
  \item  \noindent \textbf{Multitask learning in a unified framework.}
  Various facial analysis tasks are treated as distinct entities, trained and supervised individually. 
  This approach stems from the fact that task learning is closely tied to specific annotation information for real samples.
  However, when it comes to multitask learning, the inclusion of multiple annotation sources becomes a considerably more time-intensive and labor-demanding endeavor. 
  Moreover, from the perspective of model development, the incorporation of supplementary modules to cater to different subtasks presents an additional challenge.
\end{itemize} 

To address the above challenges, we propose the following strategies: 
Initially, we intend to unify face detection and FLD within a single framework, leveraging the YOLOv8 detection framework as a foundation for  learning. 
 We draw inspiration from YOLOPose's methodology, a framework tailored for human pose estimation.
 By adapting the human label into face label, we expand the scope of the human body key points to encompass a wider array of facial landmarks. This approach is intricately woven into the design of our baseline model.
 We introduce our YOLO-based model for multitask learning, aptly named YOLOMT. 

\begin{figure}[t]
  \begin{center}
      \includegraphics[width=1 \linewidth]{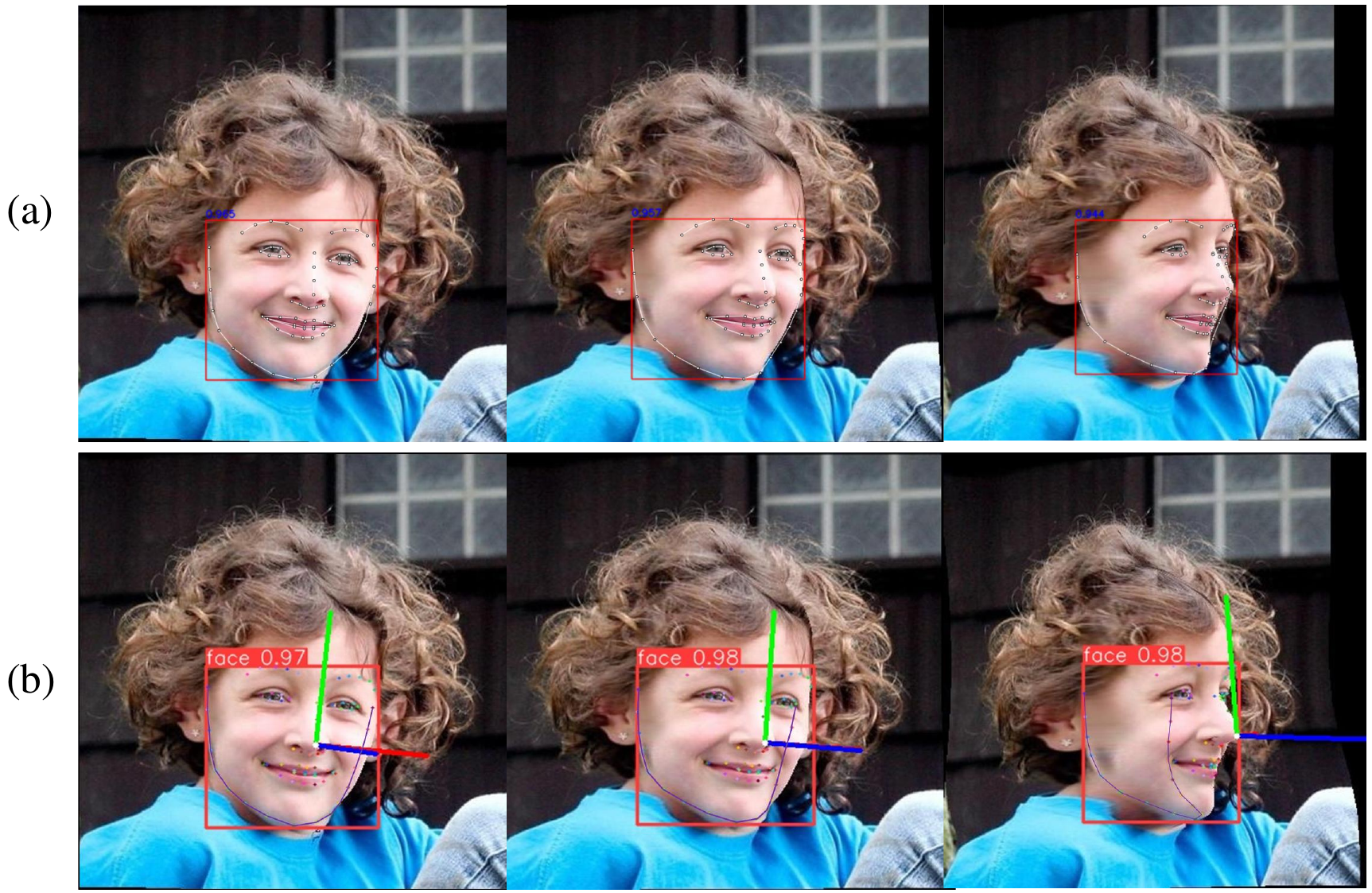}\\
      \caption{
        Distinct approaches to modeling 2D facial landmarks. (a) 2D FLD extracted from observable contours. (b) 2D FLD acquired by projecting its corresponding 3D coordinates.  (Best view in zoom).
      }
      \label{2D3DCompare}
  \end{center}  
\end{figure}

 For head pose estimation, a conventional technique involves utilizing the classic PnP algorithm. This method entails comparing the estimated 2D facial keypoints with the 3D keypoints of a standardized  model to derive a rotation matrix, subsequently yielding the three Euler angles that define the head pose. 
 However, a drawback of this method is its reliance on 2D key point estimations derived from visible contours, while actual 2D key points are obtained through the projection of 3D key points. 
 The disparity between the two can be observed in Fig. \ref{2D3DCompare}. 
 To mitigate computational overhead, we  forego designing additional models for direct head pose estimation. Instead, we opt for an indirect approach by estimating the 2D key points resulting from the 3D projection, thereby inferring the head pose indirectly.
 
 To handle the scarcity of labeled data for extreme facial poses, data augmentation and transfer learning are employed in our approach. By artificially generating data with augmented poses and leveraging pre-trained models on related tasks, the model can better generalize to challenging scenarios.

To achieve real-time performance, we base our model on the lightweight version of YOLOv8 architecture to balance  accuracy and efficiency. 
We also  undertake optimal learning  on certain original modules of YOLOv8, including reparameterized stem and bottleneck. 
We seamlessly incorporate the reparameterization technique into traditional convolutional operations, thereby transforming the original YOLO components to possess richer feature representations while maintaining a reduced computational burden.  
Empirical experimentation   demonstrates that enhancing these modules leads to an observable enhancement in the accuracy of FLD.





The contributions   of this work  are summarized as follows: 
\begin{itemize}
  \item   
  We achieve an end-to-end approach for concurrent face detection and FLD, built upon the latest YOLOv8 framework. 
  To improve the balance between the inference speed and detection accuracy, we  adopt some cheap operations such as the reparameterization technique to modify the original YOLO components (stem and bottleneck)  for  accurate FLD.

  \item   
  Different from other conventional 2D FLD approaches that primarily focus on identifying 2D key points   derived from visible contours,  
  we  can estimate   accurate  2D facial landmarks  that directly correspond to their underlying 3D structures. 
  Experimental results on the   AFLW2000-3D dataset  \cite{sagonas2016300} demonstrate that  our proposed YOLOMT can  achieve top 1 accuracy of   FLD with a mean normalized mean error (NME) of 3.02.
 
  \item 
  Capitalizing on the estimated 2D facial landmarks, we   achieve nearly effortless and accurate HPE by employing the PnP algorithm. Notably, the accomplishment of these three tasks necessitates the implementation of an end-to-end model. 
  Moreover,  experiments on  AFLW2000-3D  dataset show that 
  our tiny YOLOMT  maintains robustness and accuracy even when confronting extreme facial postures and achieving an impressive detection speed of 102 frames per second (FPS).

\end{itemize}
  

The rest of the paper is organized as follows. In Section  \ref{IOT: RelatedWork}, we provide a review of related works on multitask learning including the task of face detection, FLD  and  HPE. Section \ref{Methodology} presents the detailed description of our proposed YOLOMT. In Section \ref{IOTS: Experiments}, we compare and analyze the experimental results of our method with other existing methods. Finally,  we draw our conclusions in Section \ref{Conclusion}.

\begin{figure*}[t]
  \begin{center}
  \includegraphics[width=0.92    \textwidth]{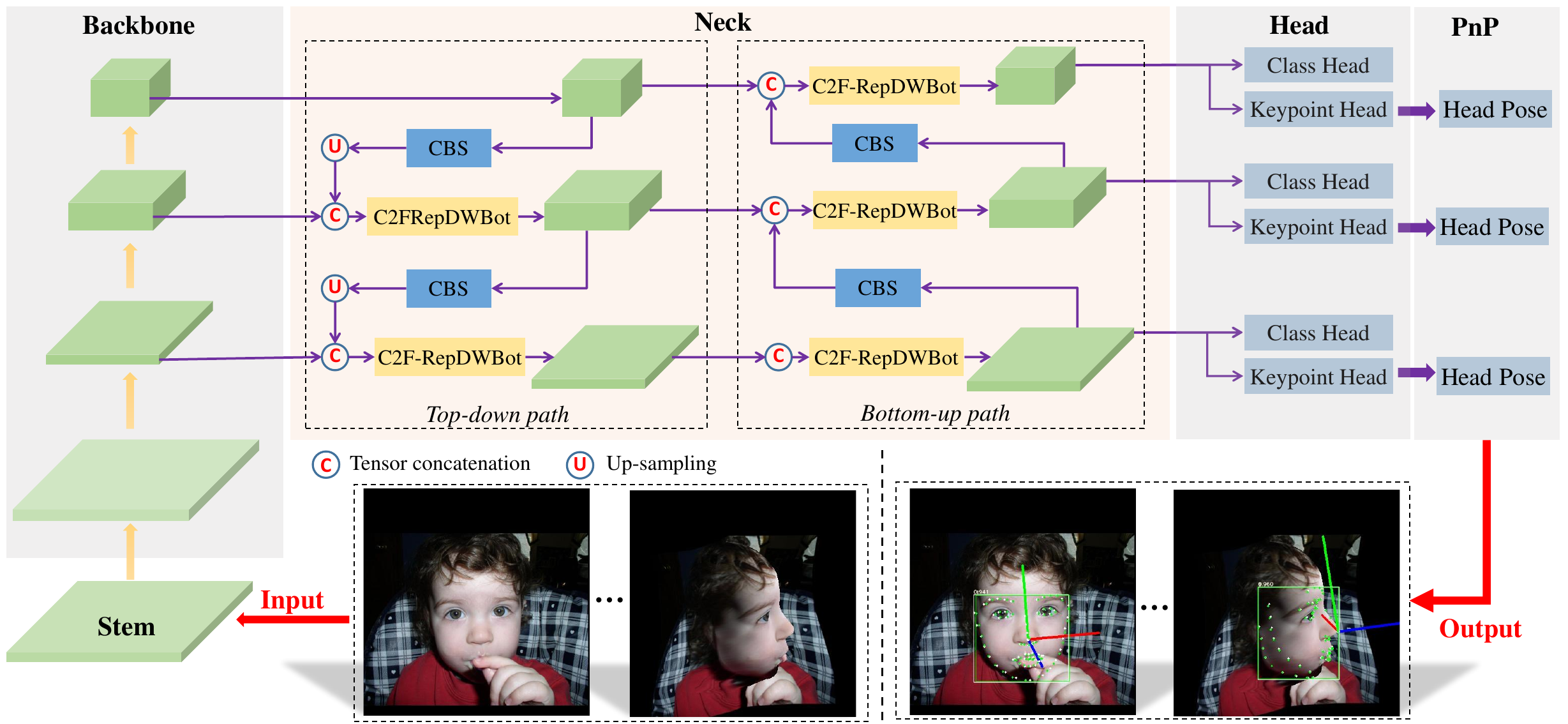}\\
  \caption{
     The overall structure of our proposed  YOLOMT.  
    It contains five components (i.e., Stem, Backbone, Neck, Head and the PnP part). 
     The Stem   is designed to generate low-semantic feature maps, serving as input to the Backbone to generate a hierarchy of  pyramid features. The Neck is designed with a bidirectional network to effectively fuse these features. 
      The Head consists of two branches:  one     for face detection and the other for FLD. Subsequently, the PnP  algorithm is employed to evaluate the head posture by leveraging the information obtained from the detected keypoints.
    }  
  \label{yoloFramework}
  \end{center}
\end{figure*}

  \section{Related Works}  
  \label{IOT: RelatedWork}

\subsection{Multitask Learning} 

In the context of learning paradigms, single-task learning (STL) involves treating each attribute or task as an isolated entity. A distinct model is created for each task, disregarding any potential connections between them. Conversely, multi-task learning (MTL) entails the development of a unified model to learn shared representations that simultaneously benefit multiple tasks.


Within the domain of facial-related detection tasks, such as  FLD  and  HPE, MTL has garnered significant adoption. It offers the advantage of concurrently detecting multiple facial attributes. 
 For example, a residual multitask learning framework is proposed in \cite{chen2021residual}  to jointly achieve FLD and expression recognition simultaneously by using the complementary information between the two tasks.
In \cite{chen2022orthogonal}, a  MTL framework is adopted in a multi-view   facial expression recognition model that regards  HPE  as an auxiliary task.
In \cite{hong2018multimodal}, a multitask manifold  method is proposed for HPE by using multimodal data.
In \cite{ranjan2017hyperface}, it proposes a multi-task learning algorithm    by fusing the intermediate layers of  deep separate CNN 
for simultaneous face detection, FLD, HPE and gender recognition.
Our proposed  YOLOMT, adapts the MTL paradigm within the YOLOv8 framework, yet with several key differentiators compared to existing strategies \cite{ranjan2017hyperface, hong2018multimodal, chen2022orthogonal, zhu2016face}:
\begin{itemize}
  \item  Unlike the MTL approach in \cite{ranjan2017hyperface}, which primarily addresses sparse (5-point)  FLD, our primary focus centers around achieving a relatively dense (68-point) FLD.
  \item Differing from  existing methods \cite{hong2018multimodal, chen2022orthogonal} that leverage    auxiliary   tasks to assist   the primary task, we aim to enhance the overall  accuracy of facial attributes by harnessing   shared deep features.
  \item Unlike the methods in \cite{guo2020towards, zhu2016face}, which adopt  a two-step pipeline involving  CNN features followed by attribute   classifiers, our proposed YOLOMT represents an end-to-end   learning approach;
\end{itemize}

  \subsection{Facial Landmark Detection with Deep Learning}

   FLD  is a prominent research area within computer vision, involving the precise localization of predefined landmarks on human faces. Over the years, FLD has evolved, with deep learning techniques leading to remarkable strides in accuracy and robustness. The methodologies in this field can be broadly categorized into two groups: regression-based and heatmap-based methods.

  Currently, the  state-of-the-art (SOTA)   performance is derived from the two-stage paradigm. This approach initially captures faces using ground truth and subsequently focuses on refining networks for precise FLD. Some implementations employ high-resolution structures to generate heatmaps for accurate landmark localization, as seen in ADA \cite{chandran2020attention}, AWing \cite{wang2019adaptive}, and HRNet \cite{wang2020deep}. For instance, ADA employs attention-based networks for high-resolution facial image predictions. AWing introduces adaptive wing loss to enhance heatmap regression, while HRNet maintains high-resolution representations throughout the process for accurate keypoint estimation.

  Addressing the complexity of high-resolution heatmap generation, the Pixel-in-Pixel Network (PIPNet)  \cite{JLS21} integrates local constraints from neighboring landmarks. Popular FLD models like   Convolutional Pose Machines (CPM) \cite{wei2016convolutional}, Hourglass Networks \cite{newell2016stacked},  DLIB \cite{king2009dlib}, 300W \cite{sagonas2016300}, and MTCNN  \cite{zhang2016joint} have been widely utilized in various applications.

  Recent trends indicate a shift towards the single-stage paradigm for FLD. In this approach, FLD is treated as an additional task during human face detection, leading to efficient detection with reduced computational demands. YOLO5Face \cite{qi2021yolo5face}, for instance, adopts an end-to-end approach for sparse (5-point) FLD, differing from our focus on dense (68-point) FLD. While RetinaFace \cite{deng2020retinaface} jointly detects faces and dense facial landmarks, its training remains unsupervised, and detailed information regarding its methodology is yet to be disclosed.  It's worth noting that achieving end-to-end dense keypoint detection is a formidable task, grappling with challenges such as training convergence and localization accuracy concerns.


\subsection{Head Pose Estimation} 

HPE addresses the task of ascertaining the three-dimensional orientation and direction of the human head, typically captured in computer vision through Euler angles. These angles encompass rotation, pitch, and roll, collectively describing the head's spatial orientation. HPE holds significance across diverse domains, finding applications in robot vision, motion tracking, and single camera calibration. 
Nevertheless, it presents inherent challenges stemming from complex backgrounds, including extreme head postures, facial expressions,   and occlusions. 

Numerous methodologies have been developed to address these challenges. 
For instance,   a matrix fisher distribution module  is proposed  in \cite{9435939} to achieve robust head pose estimation in  low pose tolerance. 
In \cite{Yang_2019_CVPR}, it introduces a lightweight model for head pose estimation from a single image,  achieved through the acquisition of a fine-grained structural mapping that spatially organizes features prior to their aggregation from a single image. 
A   nonuniform Gaussian-label   network  \cite{liu2021ngdnet}   is proposed for   HPE  under active infrared (IR) illumination. 
  \cite{Kellnhofer_2019_ICCV} presents a 3D gaze model that   learns temporal information to estimate gaze uncertainty in the wild.
This paper aims to achieve real-time joint detection while simultaneously estimating head posture using facial landmarks. It particularly focuses on scenarios involving extreme head postures, such as extreme head swings or elevated head angles.

\section{Methodology}
\label{Methodology}

 
In this section, we provide a comprehensive overview of our approach, YOLOMT. This methodology leverages the fundamental YOLOv8 framework and incorporates a dual-branch detection head. This enhanced architecture facilitates the concurrent detection of both faces and their detailed landmarks. By leveraging essential facial landmarks, particularly those associated with the eyes, mouth, nose, lower lips, and left and right cheeks, we utilize the Perspective-n-Point (PnP) algorithm to accurately estimate head postures. This approach remains robust even when addressing extreme orientations.
 

\begin{figure}[t]
  \begin{center}
    \includegraphics[width=1 \linewidth]{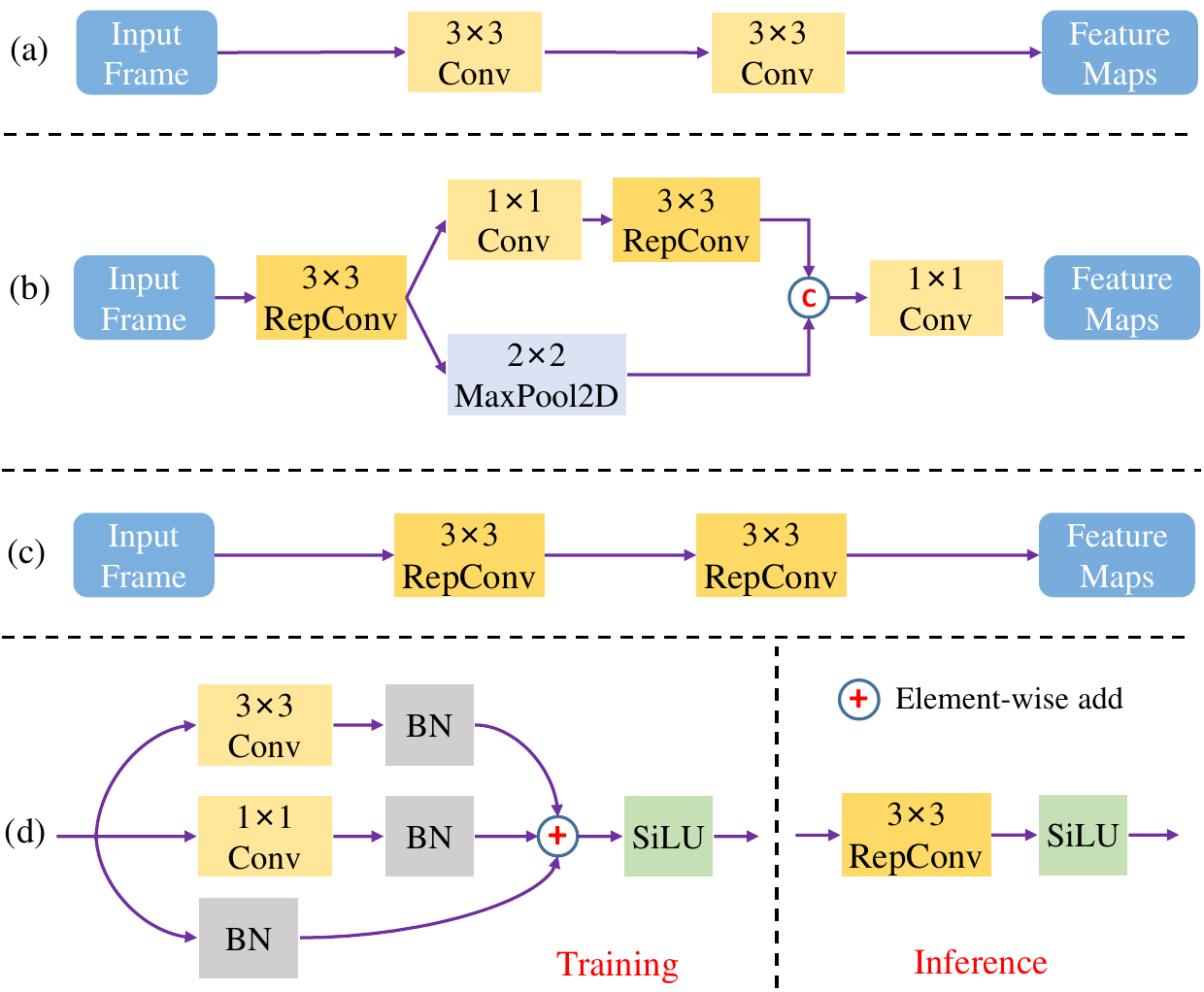}\\
    \caption{
      Various stems. 
      (a) The  naive YOLOv8 stem. 
      (b) The RepV7Stem.
      (c) The  RepV8Stem.
      (d) Different stages in RepConv.
      }
    \label{StemLearn1}
  \end{center}  
\end{figure}

\subsection{Network Architecture}

The overall architecture for FLD is illustrated in Fig. \ref{yoloFramework}, which  contains four components (i.e., Stem, Backbone, Neck and Head).

\subsubsection{\textbf{Stem}}

The initial stem module   takes the raw input image(s) as input and generates low-level semantic feature maps as output. 
 We aim to devise a resource-efficient approach for this stage, one that can capture a wealth of features while keeping computational demands low.

We explore  three stem structures, which are depicted in Fig. \ref{StemLearn1}. The original YOLOv8 stem is illustrated in Fig. \ref{StemLearn1}(a), which contains  two simple convolutional operations. In Fig. \ref{StemLearn1}(b), we present the Re-parameterized YOLOv7 stem, featuring two branches: a max-pooling branch, designed to reduce computational load, and a small-kernel branch (k = 1, 3), aimed at enhancing the visual receptive fields of the features. 
We embrace the RepConv technique to replace the original Conv3×3. 
Furthermore, we apply RepConv to directly modify the YOLOv8 stems, as depicted in Fig. \ref{StemLearn1}(c).

The details of RepConv are as shown in Fig. \ref{StemLearn1}(d), where it capitalizes on employing small-kernel Convolutions (k = 0, 1, 3) within a multi-branch architecture. This strategy facilitates the acquisition of intricate features during the training phase, after which the kernels are re-parameterized to form a singular 3×3 kernel during the inference phase. As a result, our approach can effectively learn diverse features with varying visual receptive fields during training, and streamline computations by consolidating multi-branches into a unified stream during inference. 
Both Conv and RepConv employ the Sigmoid Linear Unit (SiLU) as their activation function. 
 We refer to the    operational unit, consisting of Convolution, Batch normalization, and SiLU activation, as CBS. 
 The ablation study of  the Stem modules will be presented in Sec. \ref{IOTS: ablationStem}  within the experimental section. 

  \begin{figure}[t]
    \begin{center}
      \includegraphics[width=1 \linewidth]{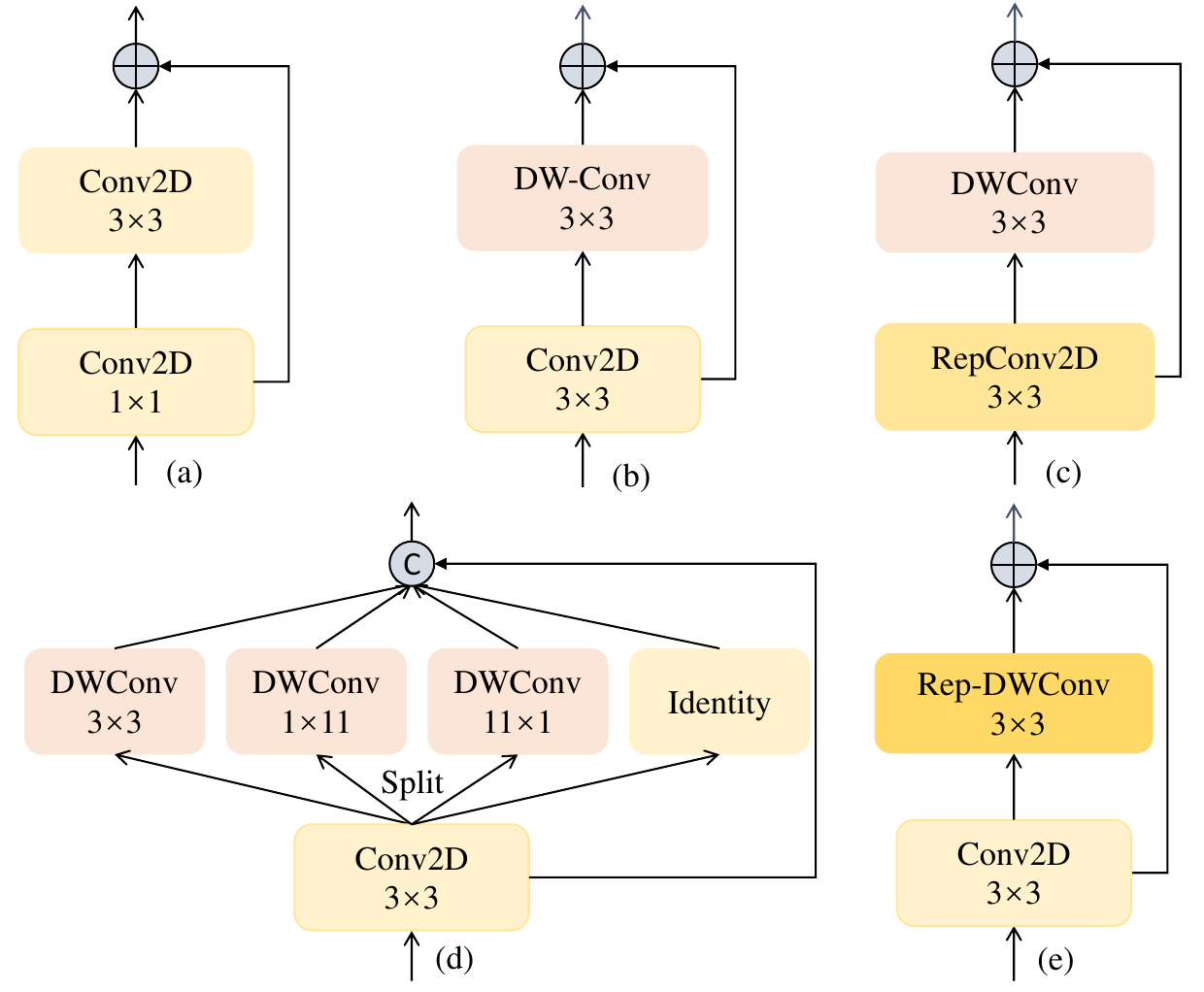}\\
      \caption{
        Various bottlenecks. 
        (a) The  YOLOv5 bottleneck. 
         (b) The YOLOv8 bottleneck. 
         (c) Reparameterized YOLOv8 bottleneck with Conv2D. 
         (d) InceptionNext  bottleneck. 
         (e) Reparameterized YOLOv8 bottleneck with DWConv2D. 
        }
      \label{BottleNeckLearn1}
    \end{center}  
  \end{figure}

  \subsubsection{\textbf{BottleNeck}}
  
  
  The backbone's fundamental unit, referred to as a bottleneck, holds a pivotal role in acquiring effective features.
   Inspired by the success of ResNet \cite{he2016deep}, contemporary   architectures incorporate residual connections to shape the bottleneck.
   In the context of YOLOv5, a simplified version of the residual bottleneck, termed V5Bot, is constructed using two convolutional layers, as depicted in Fig. \ref{BottleNeckLearn1}(a).
   Fig. \ref{BottleNeckLearn1}(b) displays the YOLOv8 bottleneck (v8Bot), wherein a larger kernel size of 3×3  is adopted in Conv and DW-Conv.

   To further amplify feature representation while conserving computational resources,
    we not only employ reparameterization to modify v8Bot but also incorporate the bottleneck concept inspired by InceptionNext\cite{liu2022convnet}.
   For instance, as depicted in Fig. \ref{BottleNeckLearn1}(c), we utilize RepConv to replace the conventional Convolutional layer in v8Bot, resulting in the formation of Repv8Bot. In Fig. \ref{BottleNeckLearn1}(e), we employ RepConv to replace the  DWConv  in v8Bot, leading to RepDWv8Bot. Additionally, Fig. \ref{BottleNeckLearn1}(d) illustrates the detailed structure   of InceptionBot.


In Sec. \ref{IOTs: ablationBot}, we will present a comprehensive ablation study that focuses on bottleneck learning. Through this study, we provide evidence that RepV8DWBot achieves a more favorable balance between accuracy and model capacity compared to the other bottleneck configurations shown in   Fig. \ref{BottleNeckLearn1}.

\subsubsection{\textbf{Neck}}
\label{IOTS: Neck describe}


The neck consolidates hierarchical features from the backbone, fostering interaction among diverse scales to enhance detection, particularly for multi-scale targets. Our neck module employs  the naive path aggregation network (PAN) with a bidirectional structure: the top-down and bottom-up paths. This dual approach results in multiscale, feature-rich outputs through reversible processes.

\begin{figure}[t]
  \begin{center}
    \includegraphics[width=1 \linewidth]{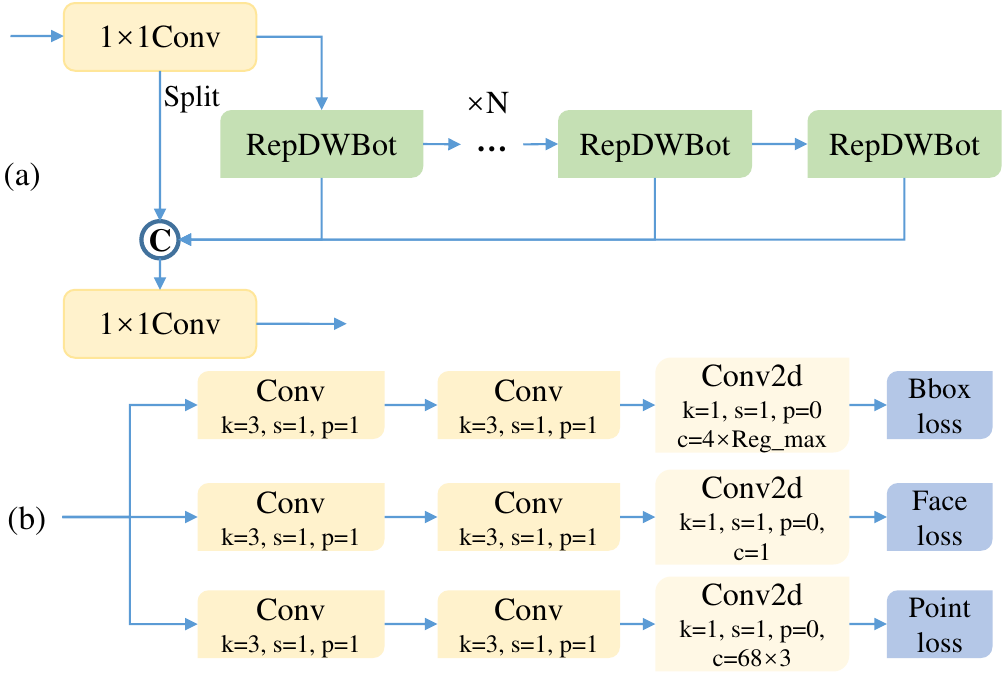}\\
    \caption{
      Different modules. (a) C2f structure with RepDWBot. (b) Detection head with three branches: one for face BBox localization, another for face label prediction, and the third one for facial keypoint localization.  
      }
    \label{C2f}
  \end{center}  
\end{figure}

We employ the proposed RepDWv8Bot to design the Cross-Stage-Partial-connections (CSP) module, a prevalent lightweight block within the YOLO framework for learning intricate gradient flow patterns. 
Named RepC2f due to its incorporation of two fast convolutional operations with RepDWBot, the detailed structure of   RepC2f   is depicted in Fig. \ref{C2f}(a).  We can see that this design   promotes a more extensive gradient flow among these cascaded bottlenecks.

   \subsubsection{\textbf{Head}}
    
 
   The detection head is designed as an adapter to generate  outputs    tailored to the specific target tasks.  
   Beyond the   naive detection head of YOLOv8 that  regresses  the   face  bounding-boxes (BBox) and predicts the face label,
     an extra branch is introduced to handle   facial landmark regression. Fig. \ref{C2f}(b) shows the   detection head with three branches: the face bbox branch, the face label branch  and the landmark branch. 
     Each of these branches consists of three consecutive convolutional operations, with the number of output channels in the last convolutional layer being specifically tailored to the corresponding task.


     

\subsubsection{\textbf{PnP}}


The Perspective-n-Point (PnP) problem entails determining the pose of a calibrated camera when provided with a set of n 3D points in the world and their corresponding 2D projections in the image. The camera's pose encompasses 6 degrees of freedom (DOF), which consist of the rotation components (roll, pitch, and yaw) and the 3D translation of the camera concerning the world. The primary objective is to find the optimal rotation and translation values that minimize the reprojection error arising from the correspondence between 3D and 2D points.

\begin{figure}[t]
	\begin{center}
		\includegraphics[width=1 \linewidth]{ 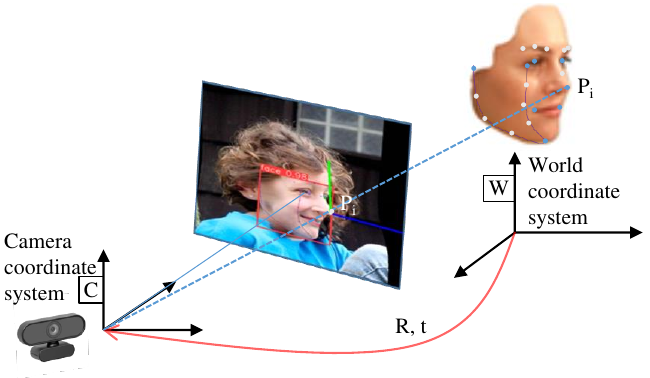}\\
		\caption{
			The PnP algorithm.  
      }
		\label{PnP}
	\end{center}  
\end{figure}

In the process, as shown in Fig. \ref{PnP}, the points initially defined in the world coordinate frame $\mathbf{X}_w$  are projected onto the image plane $[u,v]$. This projection adheres to the conventional perspective projection model used in cameras and is formulated as:
\begin{equation}
\begin{bmatrix}
  u\\ 
  v\\ 
  1
  \end{bmatrix}=\begin{bmatrix}
  f_x & 0 & c_x\\ 
  0 & f_y & c_y\\ 
  0 & 0 & 1
  \end{bmatrix} \begin{bmatrix}
  1 & 0 & 0 & 0\\ 
  0 & 1 & 0 & 0\\ 
   0& 0 & 1 & 0
  \end{bmatrix}
  \begin{bmatrix}
    \mathbf{R} & \mathbf{t}\\ 
  0 & 1
  \end{bmatrix}
  \begin{bmatrix}
  X_w\\ 
  Y_w\\ 
  Z_w\\ 
  1
  \end{bmatrix},
\end{equation}
 where  $f_x$ and $f_y$ correspond to the scaled focal lengths,   $c_x$ and $c_y$ signify the principal point on the image plane.   The variables   $\mathbf{R, t}$  
 denote the 3D rotation and 3D translation of the camera, respectively, which are the parameters being calculated (referred to as extrinsic parameters). These values are essential for characterizing the camera's pose with respect to the world.

\begin{figure}[t]
	\begin{center}
		\includegraphics[width=1 \linewidth]{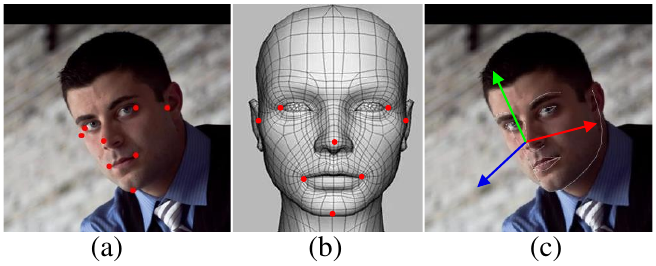}\\
		\caption{
			Selecting specific  facial keypoints to estimate the head posture by using PnP.  (a) 
      }
		\label{pnp8point1}
	\end{center}  
\end{figure}

 The Efficient PnP (EPnP) method, introduced in  \cite{lepetit2009ep}, offers a solution to the general PnP problem for cases where  n $\geq $ 4. This method operates on the premise that each of the n points (referred to as reference points) can be expressed as a weighted sum of four virtual control points. Consequently, the coordinates of these control points become the unknowns in the PnP problem, from which the final camera pose is determined.
We choose 8 estimated 2D keypoints, specifically the chin, nose tip, left and right mouth corners, left and right eye corners, and left and right cheeks, as depicted in Fig. \ref{pnp8point1}(a). By utilizing the EPnP algorithm to align these key points with their respective positions in the 3D model, as shown in Fig. \ref{pnp8point1}(b), we can  calculate the rotation matrix $\textbf{R}$ and the translation matrix $\textbf{t}$. Fig. \ref{pnp8point1}(c)   shows the head posture corresponding to the rotation matrix $\textbf{R}$.
 
\subsection{Face Detection}

Building upon \cite{maji2022yolo}, a human pose etimation framework rooted in the YOLOv5 architecture, we adapt  it to detect face label and their 68 key point positions instead of human label with sparse 17 keypoints. Unlike YOLOv5, YOLOv8 does not rely on predefined anchors for extracting candidate sets; it operates in an anchor-free manner. 
 The header module in YOLOv8   undergo  a transformation from the coupled structure of YOLOv5 to the decoupled structure depicted in Fig. \ref{C2f}(b). This modification  retains the decoupled classification and regression branches. Furthermore, the regression branch has been reformulated using the integral form proposed in the distribution focus loss (DFL).

The detection head in YOLOv8 provides two types of information for each grid: the BBox location and the face confidence. The BBox location comprises four elements, namely the box center $(C_x, C_y)$ and its width and height $(w, h)$. This information is compactly represented by a vector containing 5 elements, which is generated by the detection head for each grid. The vector  can be expressed as:
\begin{equation}
  \label{E1}
  \begin{array}{rcl}
  O_f &=& (C_x, C_y,w, h,   face_{conf}).
  \end{array}
\end{equation}

\subsection{Facial Landmark Detection}



We   extend  the original YOLOPose framework, transforming the task from regressing 17 human keypoints to regressing 68 facial landmarks. In contrast to directly detecting 2D keypoints on visible contours, our approach involves detecting 2D keypoints following a 3D projection. This method possesses the distinct advantage of preserving contour information, even in scenarios involving facial movements.

Whether it pertains to a human keypoint or a facial landmark, each point  information encompasses its coordinates and confidence level, denoted as $(L_x,L_y, conf)$. In total, the 68 facial landmarks result in 204 output elements. The landmark head in our YOLOMT, for each anchor, produces a 204-element vector, defined as:
\begin{equation}
  \label{E2}
  \begin{array}{rcl}
  O_l &=& ( L_x^{1},L_y^1,L_{conf}^1,  \dots,  L_x^{68},L_y^{68},L_{conf}^{68}). \\
  \end{array}
\end{equation}


\subsection{Loss Function}
\label{IOTS: Loss}

For the YOLOv8 model, it utilizes the Varifocal Loss (VFL) \cite{zhang2021varifocalnet} for classification and a combination of CIOU Loss \cite{zheng2020distance} and Distribution Focal Loss (DFL) \cite{li2020generalized} for regression. 

\subsubsection{Face classification loss}



The key enhancement in VFL is the introduction of asymmetric weighted operations, specifically tailored for positive and negative samples. This addresses the challenge of fluctuating weights caused by the imbalance between positive and negative samples, particularly crucial as accuracy calculations primarily focus on positive samples.
The VFL 
 is defined as follows:
\begin{equation}
  VFL(p,q) =\left\{
  \begin{array}{ll}
  -q(qlog(p)+(1-q)log(1-p))   & q>0\\
  -\alpha p^\gamma log(1-p)  & q=0,\\
  \end{array} \right.
\end{equation}
 where $p$ denotes the predicted IoU-aware classification score (IACS),  $q$ represents the target score.
  For foreground samples, their ground truth class $q$ is set as the  IoU  between the generated BBox and its corresponding ground truth. Conversely, for background samples, the target $q$ for all classes is uniformly set to 0. 
  The scaling factors  $\alpha$ and $\gamma$ are employed to  selectively reduce  the contribution of loss from negative examples 
  ($q$$=$$0$) while  preserving the weight of positive examples ($q$$>$$0$).


\subsubsection{Face BBox loss}


Advanced variant  of IoU loss like CIoU loss has become the preferred choice in many modern object detectors for bounding box tasks. 
This preference  originates from CIoU's scale-invariant properties and its direct optimization of the evaluation metric. 
In our approach, we employ the CIoU loss for bounding box supervision. For a ground truth Bbox that is associated with the grid cell $g$ located at coordinates $(x, y)$ and scale $s$, the loss can be defined as follows:
\begin{equation}
  \label{CIOU}
  \begin{array}{r}
  L_{box}(s,x,y,g) = 1-CIoU(Box_{gt}^{s,x,y,g}, Box_{pred}^{s,x,y,g}). \\
  \end{array}
\end{equation}

Furthermore,  to expedite the network's concentration on positions proximate to the target, the DFL  is introduced to enhance the network's attention towards values in close proximity to the target. DFL   optimizes the likelihood of positions both to the left and right of the target label, employing a cross-entropy formulation. This enables the network to swiftly converge towards a distribution centered around the genuine floating-point coordinates, with weights determined through linear interpolation from integer coordinates. This practice, referring to YOLOv8, employs DFL for rapid Bbox  positioning.

\subsubsection{Face keypoint loss}

Given that both keypoint detection and object detection tasks involve localization, MS COCO \cite{lin2014microsoft}   introduces the Object Keypoint Similarity (OKS) metric for assessing keypoint regression. OKS plays a role analogous to that of Intersection over Union (IoU) in object detection. 
 We adhere to the definition provided by MS COCO, which is formulated as follows:
  \begin{equation}
      OKS= {\textstyle \sum_{i}^{}} [exp(-d_{i}^{2} /2s^{2} k_{i}^{2} )\delta (v_{i} >\theta )] / {\textstyle \sum_{i}^{}[\delta (v_{i}>\theta  )]},
  \end{equation}
where $d_{i}$  represents the Euclidean distance between the predicted keypoint and the ground truth; 
$v_{i}$ stands for the visibility flag associated with the ground truth keypoint; 
$s$ denotes the object scale.
  $k_{i}$ is a per-keypoint constant that governs the rate of falloff.
  The OKS metric ranges from 0 to 1. The keypoint loss can be defined as:
  \begin{equation}
    \label{kptsLoss}
    \begin{array}{r}
    L_{kpts} = 1-\sum_{n=1}^{N_{kpts}} OKS. \\
    \end{array}
  \end{equation}

  For each individual keypoint, we train a confidence parameter that signifies the presence or absence of that particular keypoint. In this training process, the visibility flags for keypoints are utilized as the ground truth. The keypoint confidence loss can be represented using Binary Cross-Entropy (BCE), which quantifies the loss between the predicted confidence of the $n^{th}$ keypoint and its corresponding ground truth. This can be formulated as follows:
  \begin{equation}
    \label{kobjLoss}
    \begin{array}{r}
    L_{kobj}(s,x,y) =  \sum_{n=1}^{N_{kpts}}BCE(\delta(v_n>0),p^n_{kpts}). \\
    \end{array}
  \end{equation}

  Ultimately, the total loss is computed by summing across all scales, anchors, and locations:
\begin{equation}
  \label{totalloss}
  \begin{array}{rcl}
  L_{total} &=& \sum_{s,x,y,g}(\lambda_{cls}L_{vfl} + \lambda_{box}L_{box} + \lambda_{dfl}L_{dfl}   \\
      & & + \lambda_{kpts}L_{kpts} + \lambda_{kobj}L_{kobj}), \\
  \end{array}
\end{equation}
 where $\lambda_{cls}=0.5, \lambda_{box}=7.5, \lambda_{dfl}=1.5,  \lambda_{kpts}=12, \lambda_{kobj}=1.0$ are hyperparams chosen to balance between losses at different scales.
 These hyperparameter values are adopted from YOLOv8 and serve to achieve an appropriate balance among the different losses.

\section{Experiments and analysis}
\label{IOTS: Experiments}

In this section, we provide a thorough introduction of the experimental process, including   details about the datasets, the criteria used for evaluating each task,  comparisons of performance with other SOTA methods. 
     
\subsection{Datasets and Evaluation Metrics}

We proceed to evaluate the performance of our YOLOMT   for face detection and  FLD  across two established benchmark datasets: 300W-LP \cite{zhu2016face} and AFLW2000 \cite{zhu2016face}. 
Additionally, we assess the effectiveness  for  HPE  on   AFLW2000-3D \cite{zhu2016face}.

\subsubsection{300 Faces in-the-wild with large pose (300W-LP)}  


 Referred to as  ``300-W Large Pose", the 300W-LP dataset builds upon the original 300-W dataset \cite{6755925} by synthesizing  facial images that span an extensive range of head orientations, including markedly extreme poses, ranging from frontal to profile views. 
 This amalgamation results in the 300W-LP dataset, which has become a widely adopted benchmark for tasks related to facial landmark detection and pose estimation.

 We partition the dataset  into two primary subsets. The training set comprises the LFPW and Helen subsets, totaling 54,232 images. In contrast, the testing set encompasses the AFW and IBug  subsets, containing 6,993 images.

\subsubsection{AFLW2000-3D}
 
The AFLW2000-3D dataset consists of 2000 meticulously annotated images, each equipped with a comprehensive set of 68-point 3D facial landmarks at the image level. 
Due to its inherently demanding characteristics and the richness of its annotations, the AFLW2000-3D dataset has emerged as a dynamic benchmark for the evaluation of accuracy and robustness for FLD. It serves as a practical and pertinent testing ground for assessing the performance of deep learning models under conditions that closely emulate the intricacies of real-world scenarios.


 

  \subsubsection{Evaluation  Metrics}  
  Our approach encompasses the simultaneous detection of multiple tasks, including face detection, FLD and HPE.  
  These tasks are assessed using distinct evaluation metrics.
   For gauging face detection accuracy, we employ the widely-used mean average precision (mAP) strategy  \cite{lin2014microsoft} over the  IoU thresholds. 
  For evaluating the accuracy of FLD, we utilize the normalized mean error (NME) metric. This approach involves normalizing the errors by interocular distances, thus compensating for scale variations and enabling a fair comparison of results across different images.
  In the context of assessing the accuracy of HPE, we rely on the absolute mean error (AME) of Euler angles. Euler angles provide a concise representation of the head's orientation in 3D space, and the AME offers a clear measurement of the deviation between the estimated and ground truth angles.


\subsection{Experimental Settings} 
\label{IOTS: Experimental Settings}

 \subsubsection{Training} 

 As our YOLOMT framework builds upon the latest YOLOv8 detection architecture, we  strive  to maintain minimal modifications to the original YOLOv8 design. Consequently, our designed baseline  closely adheres to the configuration of YOLOv8, encompassing  training recipe including the hyperparameter settings  as depicted in Table \ref{Traing_recipe}.

 \begin{table}[t]
  \renewcommand{\arraystretch}{1.25}
  \caption{
    Training configuration.  
    }
  \label{Traing_recipe}
  \centering
  \begin{threeparttable}
  \begin{tabular}{ll|ll}
    \toprule
    Name    & Value  & Name    & Value\\ 
    \midrule
    Initial learning rate      & 0.01  &  SGD momentum      & 0.937  \\
    Warmup momentum  & 0.8 & Warmup epochs      & 3  \\
    Total training epochs      & 120 &  Weight decay      & 0.0005 \\
    Face Conf Threshold    & 0.002 & IOU Threshold    & 0.7 \\
     Batch size      & 16 & Image size      &  640×640  \\
    \bottomrule
  \end{tabular}
  \end{threeparttable}
\end{table}
  


 The initial YOLOv8 model is pretrained on the MS COCO dataset \cite{lin2014microsoft}. 
 For training our specific YOLOMT, we utilize a GPU server equipped with a GeForce 2080Ti. Each GPU has a maximum batch size of 16. We employ the SGD algorithm as our optimizer, with an initial learning rate of 1e-2. The training process is terminated at the 100-th epoch. 

   \subsubsection{ Testing}
    \label{IOTS: testing}

    During the testing phase, we maintain a consistent input resolution of 640×640 pixels. In the preprocessing step, we adhere to the standard procedure utilized in the basic YOLOv5 framework. Specifically, we resize the longer side of the input frame to the desired resolution while simultaneously applying padding to the shorter side. This procedure ensures that the resulting image is square.

For assessing the inference speed of our model, we adopt a batch size of 1. This choice allows us to evaluate the real-time performance of the model, capturing its ability to process and generate predictions swiftly.

\begin{table*}[t]
  \renewcommand{\arraystretch}{1.25}
  \caption{
    Ablation study of stems and bottlenecks.  
  }
  \label{IOTS_StemTable}
  \centering
  \begin{threeparttable}
  \begin{tabular}{l|l|cc|cccc|cccc}
    \toprule
           &             &          &           &   \multicolumn{4}{c|}{Face  }      &  \multicolumn{4}{c}{Landmark}   \\  
    Stem  & Bottleneck   & Params(M) & Flops(G) & P(\%) & R(\%) & AP@0.5  & mAP  & P(\%) & R(\%) & AP@0.5  & mAP  \\ 
    \midrule
    Naive v8Stem     & v8Bot   & 5.08      & 17.71  & 99.9 & 99.7 & 99.5 & 93.1 & 81.8 & 80.9 &  82.8  &  30.6   \\
    Rep-v7Stem       & v8Bot   & 5.08      & \textbf{17.70}  & \textbf{100.0} & 99.6 & 99.5 & 92.9 & 79.7 & 78.5 & 79.1  & 27.8  \\
    Rep-v8Stem       & v8Bot   & 5.08      & 17.71  & \textbf{100.0} &  99.8  & 99.5 & \textbf{93.3} &  83.8  &  82.8  & 82.0  & 30.1  \\
    \midrule
    Rep-v8Stem  & InceptionBot   & \textbf{4.61} & \textbf{16.73} & 99.7 & 99.4 & 99.5 & 92.3 & 78.4 & 77.7 & 78.7 & 29.7  \\
    Rep-v8Stem       & Repv8Bot    & 5.08      & 17.71  & 99.8 & 99.6 & 99.5 &  93.0  & 79.0 & 77.9 & 79.4  & 30.6  \\
    Rep-v8Stem       & RepDWv8Bot  & 5.08      & 17.71  &  99.9  & \textbf{100.0} & 99.5 & 92.7 & \textbf{84.9} & \textbf{84.1} & \textbf{85.2}  & \textbf{31.5} \\
    \bottomrule
  \end{tabular}
  \begin{tablenotes}
    \item{}   To expedite the process of ablation study, we employ  YOLOv8-mini as the foundational model.
     Bold represents the best result. 
  \end{tablenotes}
\end{threeparttable}
\end{table*}

\subsection{ Ablation Study of Stem} 
\label{IOTS: ablationStem}

Based on  the native YOLOv8 stem, we   adopt   the re-parameterization convolutions to modify the naive v8Stem to  learn feature representations  from cheap computational cost. Various stem modules have been introduced in Fig. \ref{StemLearn1}. 

 
We fully compare  their performance on 300W-LP AFLW  test set. 
As compared in Table \ref{IOTS_StemTable}, we can see that 
when v8Bot is fixed as the bottleneck:
\begin{itemize}
\item  YOLOMT with the naive v8Stem attains the highest accuracy in FLD, achieving an AP@0.5 of 82.8\% and a  mAP  of 30.6\%.
\item  YOLOMT with the naive v7Stem achieves 100\% face detection accuracy while maintaining a minimal computational cost of 17.70G. However, the FLD performance   is comparatively lower.
\item  YOLOMT with Repv8Stem achieves the highest face detection accuracy, both in terms of precision and recall, while preserving the same model capacity. Additionally, it achieves the highest precision and recall for FLD accuracy.
\end{itemize}
Based on the above comparative analysis, Repv8Stem is selected as the preferred stem module for YOLOMT.

\begin{table}[t]
  \renewcommand{\arraystretch}{1.25}
  \caption{
   Different data augmentations for training.  
  }
  \label{data_augmentations}
  \centering
  \begin{threeparttable}
  \begin{tabular}{c|cccccc}
    \toprule
   Augment      &   translate  &  scale   &   flip\_lr  &  mosaic &  mixup  & copypaste\\ 
    \midrule
    $A_0$      & 0.1 &  0.50    & 0.5 & 4 & 0.0 & 0.0 \\
    $A_1$     & 0.2 & 0.75   & 0.5 & 4 & 0.1 & 0.1\\
    $A_2$     & 0.2 & 0.90   & 0.5 & 9 & 0.1 & 0.1\\
    \bottomrule
  \end{tabular}
\end{threeparttable}
\end{table} 

\subsection{Ablation Study of BottleNeck} 
\label{IOTs: ablationBot}

The bottleneck plays a crucial role in extracting efficient and effective features. Our objective is to design an optimal bottleneck that enhances the overall performances    while preserving computational resources. We have presented various bottleneck structures in Fig. \ref{BottleNeckLearn1}.

We conducte  a performance comparison among these bottlenecks  in Table  \ref{IOTS_StemTable}. From the table, we can draw the following conclusions when Repv8Stem is selected as the stem:
\begin{itemize}
  \item  YOLOMT with the naive v8Bot achieves the highest precision and a mean Average Precision (mAP) of 93.3\% in face detection.
  \item  YOLOMT with InceptionBot has the lowest number of parameters and the least computational complexity. However, its detection metrics are not increased.
  \item  YOLOMT with Repv8DWStem achieved the highest values in five indicators, including Recall, Precision, and AP, both in face detection and FLD, while maintaining the same model capacity. 
\end{itemize}
As a result of this comprehensive evaluation, we  adopt Repv8DWStem as our preferred bottleneck.





\subsection{ Data Augmentation} 
\label{IOTS: data augmentation}


Data preprocessing have an important impact on the  model performance. In our study, we   investigate  three combinations of data augmentation strategies. As outlined in Table \ref{data_augmentations}, the original  $A_0$ augmentations, including translation, scaling, flipping, and mosaics, are applied to each frame. Specifically, image translation varies within a range of ±10\%, scale ranges within ±50\%, and the probability of left-right flipping is set at 50\%. Additionally, we integrate 4-image mosaics with a 100\% probability.


\begin{table}[t]
  \renewcommand{\arraystretch}{1.25}
  \caption{
    Ablation study of data augmentation strategies.  
  }
  \label{AugmentationStudy}
  \centering
  \begin{threeparttable}
  \begin{tabular}{c|ccc|ccc}
    \toprule
    &     \multicolumn{3}{c|}{Face  }      &  \multicolumn{3}{c}{Landmark}   \\  
    Augment & P(\%) & R(\%)  & mAP & P(\%) & R(\%)   & mAP   \\ 
    \midrule
    $A_0$  & \textbf{100.0} &  \textbf{100.0} & \textbf{98.4} &98.0 & 97.6  & 69.1   \\
    $A_1$  & 99.8 & 99.7 & 98.3 & \textbf{98.2} & \textbf{97.8} & \textbf{69.9} \\
    $A_2$  &  99.7  & 99.8 & 97.7  & 97.9 & 97.5 & 66.4\\
    \bottomrule
  \end{tabular}
  \begin{tablenotes}
    \item{}   Here, we employ YOLOv8-small as the basic  model. 
  \end{tablenotes}
\end{threeparttable}
\end{table}

\begin{table*}[t]
  \renewcommand{\arraystretch}{1.25}
  \caption{
    Performance comparison with other advanced methods on the AFLW2000-3D test set. 
  }
  \label{300WComparision}
  \centering
  \begin{threeparttable} 
  \begin{tabular}{l|l|c|rr|cc|cc|cccc}
    \toprule
    &    &     &  Params   &   Flops & \multicolumn{2}{c|}{Face}   & \multicolumn{2}{c|}{Landmark}   &  \multicolumn{4}{c}{NME}        \\  
    Method & Backbone  &  Input Size &   (M)  &  (G) &   P(\%) & mAP &   P(\%) & mAP   &  [0,30] &  [30,60] &  [60,90] & \textbf{Mean}    \\ 
    \midrule
    \multicolumn{13}{c}{Two-stage paradigm}                                                             \\ 
      \midrule
      DenseCorr  \cite{yu2017learning} & Hourglass &  128$\times$128  & -  & - & \multicolumn{2}{c|}{GT} &- &- & 3.62 & 6.06 & 9.56 & 6.41  \\
      3DSTN  \cite{bhagavatula2017faster} & VGG-16 &  250$\times$250   & 138.36 & 30.84 & \multicolumn{2}{c|}{GT} &- &- &  3.55 & 3.92 & 5.21 & 4.23     \\
      3D-FAN  \cite{bulat2017far} & Hourglass  &   128$\times$128   & 24.00  & -  & \multicolumn{2}{c|}{GT} &- &- &  3.15 & 3.53 & 4.60 & 3.76 \\
      3DDFA  \cite{zhu2017face} & - &  200$\times$200   & - & -     & \multicolumn{2}{c|}{GT} &- &- &  2.84 & 3.57 & 4.96 & 3.79   \\
      PRNet  \cite{feng2018joint} & Hourglass &   256$\times$256   & -  & -       & \multicolumn{2}{c|}{GT} &- &- &  2.75 & 3.51 & 4.61 & 3.62 \\
      2DASL  \cite{9091237} & ResNet-50 &  120$\times$120   & 23.52 & 8.79  & \multicolumn{2}{c|}{GT} &- &- &  2.75 & 3.46 & 4.45 & 3.55   \\
      3DDFAv2(MR) \cite{guo2020towards} & MobileNet &  120$\times$120 & 3.27  & 0.37 & \multicolumn{2}{c|}{GT} &- &- &  2.75 & 3.49 & 4.53 & 3.59 \\
      3DDFAv2(MRS) \cite{guo2020towards} & MobileNet &  120$\times$120   & 3.27  & 0.37 & \multicolumn{2}{c|}{GT} &- &- &  2.63 & 3.42 & 4.48 & 3.51   \\
      SynergyNet \cite{wu2021synergy} & MobileNet &  120$\times$120   & 3.80  & 0.37 & \multicolumn{2}{c|}{GT} &- &- &  2.65 & \textbf{3.30} & \textbf{4.24} & 3.41  \\
      \midrule
   \multicolumn{12}{c}{End-to-end paradigm}                                                                \\ 
   \midrule
   \qquad YOLOMT-t-$A_0$ & RepDWv8Bot &  640$\times$640 & 5.08  & 17.71 & 99.1 & 84.4 &63.8& 15.7 &  2.66 &   3.57 &  4.96 &  3.18  \\
    Our  YOLOMT-s-$A_0$ & RepDWv8Bot&  640$\times$640 & 14.02  & 40.70 & 99.3 & 84.9 & 64.8 & 15.9  & 2.57 & 3.52&4.87 & 3.11 \\
   \qquad   YOLOMT-s-$A_1$ & RepDWv8Bot&  640$\times$640 & 14.02 & 40.70 & 98.7 & 84.5 & 66.8 & 14.3  & \textbf{2.51} & 3.43 & 4.65 & \textbf{3.02} \\
   \bottomrule
  \end{tabular}
  \begin{tablenotes}
    \item It should be stated that these two-stage methods use ground-truth to extract face and do not consider the computational cost in the face-detection stage. 
     `-' indicates null or not given.  
     $t, s$   refer to the tiny and small versions of YOLOv8. $A_0, A_1$ represent the specific augmentation strategies.
  \end{tablenotes}
\end{threeparttable}
\end{table*}

\begin{figure}[t]
  \begin{center}
    \includegraphics[width=0.9 \linewidth]{ 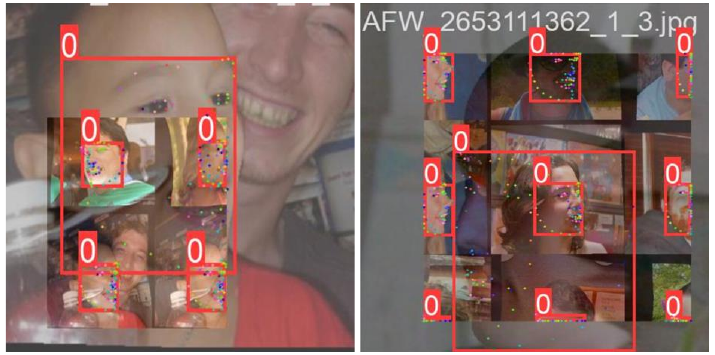}\\
    \caption{
      Different augmentation effects. 
      (a) Using $A_1$ to augment images. 
      (b) Using $A_2$ to augment images. 
      }
    \label{augment}
  \end{center}  
\end{figure}

We try to conduct comparative experiments aimed at identifying a tailored augmentation strategy that can enhance the model's capacity to handle diverse facial positions, thereby aligning with the challenges encountered in real-world detection scenarios. 
To achieve this, we   explore  two other sets of strategies denoted as  $A_1$ and $A_2$ in Table \ref{data_augmentations}.   $A_1$ places a stronger emphasis on a larger scaling ratio (0.75) and incorporates mixup and copy-paste operations. On the other hand,  $A_2$ employs even more robust enhancements, featuring an even larger scaling ratio (e.g., 0.9) and the use of 9-image mosaics. The distinct effects of these image augmentation strategies are illustrated in Fig. \ref{augment}.

Based on the results presented in Table \ref{AugmentationStudy}, the following observations can be made:
\begin{itemize}
  \item When employing the original $A_0$ data augmentation strategy, the most favorable results are obtained across all three face detection metrics.
  \item The enhanced $A_1$ data augmentation strategy, while leading to a slight reduction in face detection accuracy compared to $A_0$, achieves the highest accuracy in FLD. 
  \item The most robust $A_2$ data augmentation strategy does not yield performance improvements in either face detection or FLD.
\end{itemize}
In conclusion, the utilization of the $A_1$ data augmentation strategy enriches detection scenarios, enhancing overall robustness and keypoint accuracy. However, it comes at the expense of a minor decrease in face detection performance.

  \subsection {FLD Performance Comparisons with SOTAs}  
  \label{IOTS: performanceComparison300W}
  

  We   conduct  a comprehensive comparison with several  SOTA methods, as summarized in Table  \ref{300WComparision} . It is important to note that the current SOTA methods primarily employ a two-stage approach to achieve high-precision keypoint detection. 
  The ``GT" notation in the table indicates that they utilize ground truth   to obtain face BBoxes. Their primary evaluation metric for  FLD  is the Normalized Mean Error (NME).

  When comparing   these two-stage SOTA methods, we can observe  that both SynergyNet \cite{wu2021synergy} and 3DDFAv2 \cite{guo2020towards} utilize the  lightweight MobileNet as their backbone network, which has the smallest model capacity (i.e., the lowest number of parameters and computational cost). Impressively, they achieve high   accuracy  in FLD on the AFLW2000-3D dataset, 
  with a  mean NME approaching 3.40. Furthermore, these methods demonstrate robustness even when dealing with faces rotated at large angles (i.e., between 60 and 90 degrees).


  The AFLW2000-3D dataset presents a highly challenging scenario,  where a limited number of  images encompass more than two faces or suffer from missed detections, the potential for significant errors arises when attempting to match a detected face with its corresponding ground truth face bounding box, resulting in elevated NMEs. In line with established best practices   in PRNet \cite{feng2018joint}, we judiciously omit  the most challenging 17 cases, where the NME exceeds 20.0. Our evaluation is therefore focused on the remaining 1983 images, which are more representative of typical scenarios.
  


  Our end-to-end approach can achieve  simultaneous detection of faces and their key points, leveraging various data augmentation techniques to enhance system robustness. A comparison in Table \ref{300WComparision} reveals the following insights:
  \begin{itemize}
    \item   Our YOLOMT-s outperforms other SOTA two-stage methods when using $A_0, A_1$ data augmentation strategies, achieving the second-lowest mean NME of 3.11 and the lowest mean NME of 3.02, respectively. This demonstrates that our end-to-end approach excels in accurately locating the top 1 facial landmark on the dataset.
    \item  Within the angle range of 0 to 30 degrees, both models also exhibit top-tier localization accuracy.
    \item  In the broader ranges of 30 to 60 and 60 to 90 degrees, our approach exhibits slightly lower accuracy compared to   the SOTA SynergyNet, achieving NMEs of 3.43 and 4.65, respectively. This indicates that there is potential for improvement in our method to accurately localize facial keypoints in larger range of facial oscillations. 
    \item Our YOLOMT has a lower model capacity when using Tiny YOLOv8 as the basic framework. Compared to the small version, the parameter and computational complexity are greatly reduced, but the FLD accuracy is also slightly reduced.
  \end{itemize}

    \begin{figure*}[t]
      \begin{center}
        \includegraphics[width=1 \linewidth]{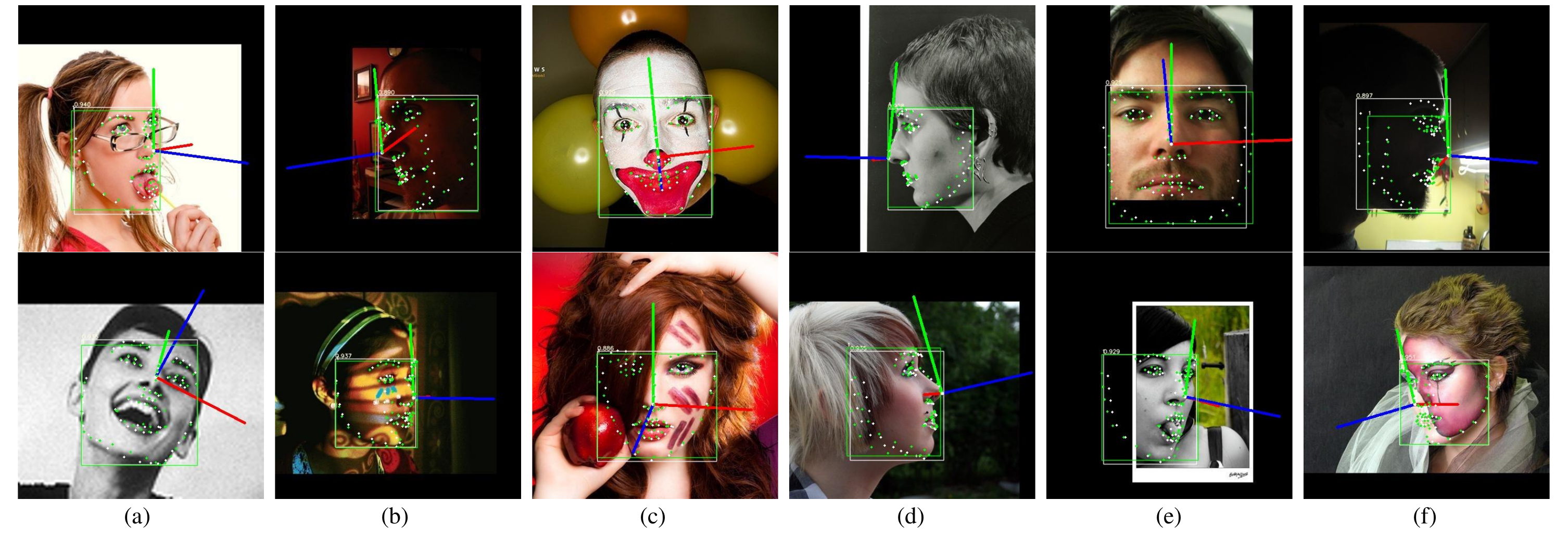}\\
        \caption{ 
          More displays of   three joint detection tasks under challenging conditions, i.e., (a) expressions, (b) low lighting,  (c) makeups,   (d) large swing angles, (e) occlusions and (f) mixed complex scenes. Green color represents ground truth and white color represents the predicted outputs. (Best view  in zoom).
        }
        \label{demo}
      \end{center}
    \end{figure*}

    \subsection{ Robust Face  Detection} 
    \label{ablationTransfer}
    
    The predominant approach in two-stage  SOTA  methods involves the dependency on  GT to  extract  faces. This strategy offers a twofold advantage: firstly, it mitigates potential accuracy loss stemming from missed or false detections. Any such misses in the initial detection stage render the subsequent stage incapable of locating crucial facial landmarks. Secondly, this approach mitigates the additional time and computational resources required for the first stage detector. It's imperative to consider both the computational cost and time associated with the first stage detector.

    However, practical applications often involve scenarios where ground truth data is unavailable. In such cases, advanced or high-speed facial detectors are typically employed in the first stage to extract facial regions. For instance, the 3DFFA method utilizes the face detector from Dlib, while 3DFFA2 adopts FaceBox \cite{zhang2017faceboxes} as its facial detection mechanism. To address this real-world scenario, we conducted empirical assessments using the AFLW2000-3D dataset.

    In our study, we perform  a comprehensive comparative analysis, evaluating the recognition accuracy of these two facial detectors in comparison to our end-to-end facial recognition system, as detailed in Table \ref{FacePrecision}.
    The outcomes of our analysis indisputably highlight the superior performance of our face detectors when compared to Dlib and FaceBox. 
    Additionally, we provide visual examples of several challenging scenarios where our detector accurately identifies faces in  Fig. \ref{demo}.

    \begin{table}[t]
      \renewcommand{\arraystretch}{1.25}
      \caption{
    Face detection performance compared with other methods.  }
      \label{FacePrecision}
      \centering
      \begin{threeparttable}
      \begin{tabular}{l|cc|cc}
        \toprule
        Method   &   DLib \cite{king2009dlib}   &  FaceBox \cite{zhu2017face} &      Our-t &      Our-s  \\ 
        \midrule
          P(\%)  &  $<$80.0 & 98.3  & 99.1 &99.3  \\
        \bottomrule
      \end{tabular}
      \footnotesize  Precision is calculated using AFLW2000-3D.
    \end{threeparttable}
    \end{table}

\subsection{Real-time Inference Speed} 
\label{IOTS: inferenceSpeed}

In practical applications, especially in real-time systems like facial landmark-based driver drowsiness detection, achieving high-speed performance is crucial. We conduct  inference experiments on a workstation equipped with a CPU (Intel(R) Xeon(R) Gold 6134 @ 3.20GHz) and a GeForce 2080Ti GPU. The inference speed is calculated by testing validation samples at a resolution of 640×640 with a batch size of 1.

\begin{table}[t]
  \renewcommand{\arraystretch}{1.25}
  \caption{Inference speed comparison with other methods.  }
  \label{SpeedTable}
  \centering
  \begin{threeparttable}
  \begin{tabular}{l|c|rr|cr}
    \toprule
    Method   & Resolution & Params   &  GFlops & NME &   FPS \\ 
    \midrule
      \multicolumn{6}{c}{Two-stage paradigm}   \\ 
      \midrule
      DAC-CSR \cite{feng2017dynamic} & 100$\times$100 &   -  & -  & 6.03  & 10 \\
      HRNet \cite{wang2020deep}    &  256$\times$256 & 9.70M    &  4.80  &  3.45 & 12 \\
      3DDFA  \cite{zhu2017face}  & 200$\times$200 & -    &  - &  3.79  & 43 \\
      3DDFAv2/FB   \cite{guo2020towards}  & 120$\times$120 & 3.27M    &  0.37 &  3.51  &  99 \\
      \midrule
      \multicolumn{6}{c}{End-to-end paradigm}   \\ 
      \midrule
      Our YOLOMT-t  & 640$\times$640 &  5.08M  & 17.71  &  3.18  &  \textbf{102}    \\
      Our YOLOMT-s   & 640$\times$640 & 14.02M  &  40.70 & \textbf{3.02} &  94   \\
    \bottomrule
  \end{tabular}
\end{threeparttable}
\end{table}

Our proposed YOLOMT   meets the real-time requirement. 
The time consumption of YOLOMT mainly consists of two processes: inference and non-maximum suppression (NMS), which take 9.6 and 0.6 milliseconds respectively. As a result, we achieve  a real-time test speed of 98 frames per second (FPS).  When utilizing the tiny YOLOMT, it achieves a speed of 102  FPS with less computational cost.
Comparing the inference speed and FLD accuracy as shown in Fig. \ref{SpeedTable} , we can see that:
\begin{itemize}
\item When our YOLOMT-s achieve  the highest accuracy in FLD (i.e., with the lowest NME of 3.02), it also achieves a  real-time detection speed of 102 FPS.
\item Our tiny YOLOMT can further improve the inference speed to 110 FPS, but there is a slight increase in NME.
\item When considering FaceBox as the face detector in this two-stage system, 3DDFAv2 can  detect VGA image at a speed of 99 FPS. More,  this speed is achieved by cropping the face  image block and   resizing it to a small resolution of 120$\times$120. 
\end{itemize}
In summary, our end-to-end YOLOMT effectively balances detection accuracy and speed, providing a well-rounded solution.
 Fig. \ref{demo} illustrates that our YOLOMT performs well in challenging scenarios involving viewpoint variations, occlusions, changes in appearance, makeup, and multi-scale conditions, among others.

\section{Conclusion}
\label{Conclusion}

In this paper, we introduce a real-time multitasking learning system named YOLOMT, built upon the YOLOv8 detection framework. YOLOMT is designed for simultaneous face detection and FLD. Furthermore, we leverage precise facial key points to estimate head posture using the PNP matching algorithm.
To enhance the trade-off between speed and accuracy, we incorporate re-parameterization  to modify the original  YOLOv8 components (i.e., Stem and Bottleneck), resulting in an improved YOLOMT system. We empirically demonstrate the effectiveness and efficiency of YOLOMT using publicly available datasets, including 300W-LP and AFLW2000, validating its performance in real-world scenarios.

 

\ifCLASSOPTIONcaptionsoff
  \newpage
\fi

\ifCLASSOPTIONcaptionsoff
  \newpage
\fi

\bibliography{ref}
\bibliographystyle{IEEEtran}

\end{document}